# Automating Computer Bottleneck Detection with Belief Nets

**John S. Breese** **Russ Blake**
Microsoft Research
One Microsoft Way
Redmond, WA, 98052-6399
(breese@microsoft.com,russbl@microsoft.com)

## Abstract

We describe an application of belief networks to the diagnosis of bottlenecks in computer systems. The technique relies on a high-level functional model of the interaction between application workloads, the Windows NT operating system, and system hardware. Given a workload description, the model predicts the values of observable system counters available from the Windows NT performance monitoring tool. Uncertainty in workloads, predictions, and counter values are characterized with Gaussian distributions. During diagnostic inference, we use observed performance monitor values to find the most probable assignment to the workload parameters.

In this paper we provide some background on automated bottleneck detection, describe the structure of the system model, and discuss empirical procedures for model calibration and verification. Part of the calibration process includes generating a dataset to estimate a multivariate Gaussian error model. Initial results in diagnosing bottlenecks are presented.

## 1 Introduction

Improving the performance of computer programs is a central theme in computer science. Considerable theoretical research (e.g. complexity analysis) and applied development (e.g. optimizing compilers) have been focused on improving performance, typically considering worst-case conditions in completely specified computing environments. However, the actual, real-world performance of a program is inherently uncertain. We typically do not know the user's patterns of access of the program's functionality. We do not know the specific performance characteristics of the software and hardware on the particular machine in question. And typically we do not have perfect measurements of internal system states; rather there are some limited set of metered outputs from which we can determine performance. In general, there has been relatively little work addressing the uncertain aspects of performance analysis in real-world, dynamic environments.

In this paper, we use the formalism of belief networks and probability as a framework for modeling the various types of uncertainty inherent in performance analysis, in particular the detection of bottlenecks. Most previous diagnostic applications have used belief networks consisting entirely of discrete variables (e.g. [Heckerman et al., 1992]). There also have been systems developed that address interpreting results from a functional analysis in a discrete belief network [Breese et al., 1992]. Recently, there have been integrated continuous/discrete approaches [Olesen, 1993], under the restriction that in the continuous portion of the network the variables have a linear-additive joint Gaussian distribution. In this paper, we develop a belief network that encodes a functional model of a computer operating system. The inference method combines expert assessments (the priors over hypotheses and the model structure) with an empirically estimated error model. Since we make no assumptions about the nature of the underlying causal model, inference provides an estimate of the most probable explanation, not a complete posterior distribution.

In Sections 2 and 3, we discuss fundamental issues in performance analysis and bottleneck detection for those readers unfamiliar with the domain. In Section 4, we present issues in developing, calibrating, and verifying the model of the operating system. In Section 5, we present methods for inferring system behavior from observables, including learning an error model for model predictions, and in Section 6 present empirical results. Eventual applications of this diagnostic capability include support for hardware purchase decision making, advanced software development tools, and dynamic system tuning.

## 2 Performance Analysis

The literature on computer performance evaluation has been largely concerned with answering the following question: Given a planned workload, what selections of possible computer equipment, interconnection schemes, protocols, and algorithms should be made to produce satisfactory performance? This problem is generally attacked by determining the relevant characteristics of the workload to be applied and the relevant characteristics of the computer system performance behavior. A model of the proposed system is constructed, verified against either the actual equipment or a simulation, and used to predict the ability of the system to handle the proposed workload [Dowdy, 1989].

Several forms of evaluation have been proposed, each with varying tradeoffs. Analytic models, typically queuing theoretic models, are inexpensive to run and permit lots of experimental explorations. However, they often must make mathematical assumptions which do not reflect real system constraints ([Schweitzer et al., 1993] and its references). Discrete event simulation models (e.g. [Yu et al., 1985]) match the internal structure and workflow algorithms of the proposed system in intricate detail, but precisely due to their detail are difficult to construct and expensive to run, reducing the number of experimental questions which can be asked. Nonetheless they provide interesting information difficult to obtain otherwise, such as the marginal distribution of response times.

Models have also been built using Operational Analysis, a measurement-based technique pioneered by Buzen. [Buzen, 1976, Buzen and Shum, 1987]. Operational models are simple and easily verifiable on data collected during any specified time interval. Operational models rely on long run averages and therefore lack detail provided by queuing and simulation models, but provide indispensable first-order (i.e., average) performance metrics.

*Atomic models* [Blake, 1979, Gray, 1987] are a refinement of operational models which assign atomic performance values to fundamental system operations. Operational service times at devices can be thought of as molecules of device consumption; atomic models depict these molecules as being made up of atoms of device consumption combined to reflect the structure of the operating system. Atomic models permit detailed operational models to be constructed and verified for existing or proposed systems, are computationally simple, but like operational models provide only average performance metrics.

All of these types of models need a description of the workload which will be applied to the systems by the anticipated application. Application workloads are naturally expressed in terms of use of the facilities the application makes available (e.g., open a document or print a report.) When possible, the original workload used to drive the model can be taken from traces of real system activity on a pre-existing system [Yu et al., 1985]. Methods have been developed for hierarchically decomposing application functional actions into system level calls for input into models [Graf, 1987, Raghavan et al., 1987]. All of this literature assumes the workload is known; the problem is in specifying it economically at the right level of detail for consumption by the model.

Over the past decade there has been a shift away from monolithic applications on dedicated hardware to a client/server environment where inter-communicating applications are thrown together on a mix of servers. Ferrari has noted that "no systematic methodologies are known to reduce a multi-computer workload's description to a more compact and representative model of that workload" [Ferrari, 1989]. The independent construction of the applications and operating systems–combined with their inter-dependent operation–make it very difficult, if not impossible, to specify the workload. Furthermore, much of the workload applied to a system may not be the direct plan of the application designer at all. For example the amount of paging traffic induced by placing a particular application onto a particular system is seldom the intent of the application designer, since typically the application co-exists on the system with other applications, and the cumulative effect determines paging behavior. These very real concerns threaten to render the large body of effort thus far expended on computer performance modeling irrelevant, because the workload to be supplied to any model cannot be known a priori.

The thrust of our research is to overcome this fundamental problem by using probabilistic methods to infer the most probable existing workload from the performance measures provided by an existing system. In the situation where a new application is to be added to an existing configuration, the inferred workload can then be combined algebraically with the anticipated application workload derived using the conventional techniques referenced above. If the interprocess interactions are too strong for the anticipated application workload to be characterized in isolation [Ferrari, 1989], a prototype of the new application embedded in the expected environment can be measured to infer the new composite workload. The resulting workload can be used to drive the many sorts of mod-els discussed above, enabling classical analyses such as identifying the bottleneck and its causes, predicting the effect of equipment purchases, or predicting the effects of changes in operating system algorithms.

Previous attempts to automate bottleneck detection expertise have been constructed using rule-based techniques and have focused on performance tuning [Irgon et al., 1988, Domanski, 1989]. These methods have no explicit representation of a workload or a model of the system. While shown to be useful for automating tuning, a rule-based approach cannot be easily manipulated to extrapolate changes in the workload, differences in hardware configuration, or revisions in operating system algorithms. In addition these methods provide no explicit methodology for managing uncertainty in the heuristics or their mapping to the rules.

## 3 Bottleneck Detection

Computer performance bottlenecks are typified by the overconsumption of some hardware resource. Usually this results in the underconsumption of other hardware resources resulting in a delay completing the workload. Once a particular resource is identified as the bottleneck, a number of remedies exist. These include distributing the load across additional instances of that resource, installing a faster resource, or redesigning the workload to use another resource instead. These actions will resolve the bottleneck by reducing the time spent using the bottlenecking resource, possibly even shifting the bottleneck to another component. Bottlenecks cannot be eliminated, only moved, because there is always some resource which can be faster to the benefit of the workload's completion time.

A *transaction* is a unit of work on a computer system. The notion of a transaction is meant in the widest sense of an interaction with the system at a level of abstraction convenient to the workload: saving a file, sending a piece of e-mail, or compiling a program. The total time the transaction uses on each system resource is called the *demand* for that resource. Based on these sorts of fundamental notions, we can characterize bottlenecks using well established Operational Laws of Computer System Performance [Buzen, 1976].

Let $D_i$ be the demand for resource $i$ and $D_j$ be the demand for resource $j$. The Consistency Law states:

$$\frac{U_i}{U_j} = \frac{D_i}{D_j}$$

where $U_i$ is the *utilization* of resource $i$. Utilization is the proportion of time that a resource is actually in use. This tells us that the devices will be busy in relation to the demand for them.

The *throughput* of a device measures the number of transactions per second a resource or system can service. One consequence of the Consistency Law is that resource utilization may not be maximum in order for a system to be achieving maximum throughput, $T_m$, defined in units of transactions/second. The maximum throughput for any resource $i$ is

$$T_i^m = \frac{1}{D_i}$$

Clearly, the resource with the smallest $T^m$ in the system for this transaction will determine the maximum throughput the system can achieve. This resource is the bottleneck. Making any other resource faster can never yield more throughput, it can only make the incorrectly improved resource have lower utilization.

For example suppose that a transaction requires 0.3 seconds of processor time and 0.5 seconds of disk time, and no other resource time in a single processor, single disk system. The processor can handle 3.3 transactions/second, while the disk can handle 2 transactions/second. So the overall system can handle only 2 transactions per second, at which point the disk will be saturated with utilization = 1. By the Consistency Law, the utilization of the processor at that point will be 0.3/0.5 = 0.6, or 60%. This gives rise to what is known as the Throughput Law, which says that for all devices, the overall throughput of the system, $T$, is given by the following:

$$T = \frac{U_i}{D_i}$$

Several problems arise which prevent the simplistic detection of bottlenecks by merely observing device utilization. The first problem is one of inadequate system instrumentation. Frequently, resource utilizations are not measured. This is partly due to a lack of fast, inexpensive, accurate clocks for timing the usage of resources, and partly due to a lack of computer industry coordination concerning the metering of resource activity and access to that information throughout the hardware hierarchy. Even in a modern system such as Microsoft's Windows NT which supports over 500 different performance metrics [Blake, 1995], inadequate instrumentation remains an impediment to bottleneck detection. Gradual improvement is being made in these areas, but for the foreseeable future there is often a need to infer device utilization indirectly.

Another problem confounding simple bottleneck detection is that certain resources are used to satisfy fundamentally different workload requirements. In the above example we deduced that the disk was the bottleneck, but we cannot simply conclude that a faster disk is the correct solution. Modern computer systems use disks and local area networks for both virtual memory and file storage. A shortage of RAM can cause disk activity as easily as file activity can, so the correct solution might be to buy more RAM, not faster disks. Even if the activity is simple file activity, modern systems also use RAM to cache file data, so even if the activity is pure file access the addition of RAM may still be the right solution. Conversely if the activity is one-time sequential file access, it is unlikely that additional RAM will be of assistance, and a faster disk is required.

Which of these causes of the bottleneck prevails in a given case is key to the correct remedy. To answer this question we must infer the bottleneck's cause from existing system metrics, an inherently uncertain endeavor. That is the intent of this research.

## 4 A Model of Computer System Performance

The system model we have developed combines workload attributes with calibrated operating system characteristics and calibrated hardware resources to predict performance monitor counters. To accomplish this it must approximate the algorithm the operating system uses to allocate the workloads to resources. Despite the wealth of services offered by modern operating systems, such a model can be built at a fairly high level of abstraction with reasonable accuracy [Blake, 1979, Gray, 1987].

A high level of abstraction is possible because a surprisingly small number of basic operations comprise the majority of sustained activity inherent in most bottlenecks. This is in part due to the kernel [Ritchie and Thompson, 1974] and microkernel [Accetta et al., 1986] approaches to operating system construction. In this design paradigm, shared in large part by the Windows NT operating system [Custer, 1993], system services are built on top of a relatively small number of primitive functions comprising the kernel of the operating system. Knowing the instruction path lengths of those kernel primitives at the root of sustained operations is sufficient to characterize the majority of system activity, since most non-primitive services are constructed using these atomic services as building blocks.

In such an atomic model of an operating system, the workload is most naturally specified at the level of the important subset of calls to invoke operating system services. Such simplifications, while providing concise, powerful models, contribute to an underlying uncertainty that the abstract model will accurately match actual system behavior.

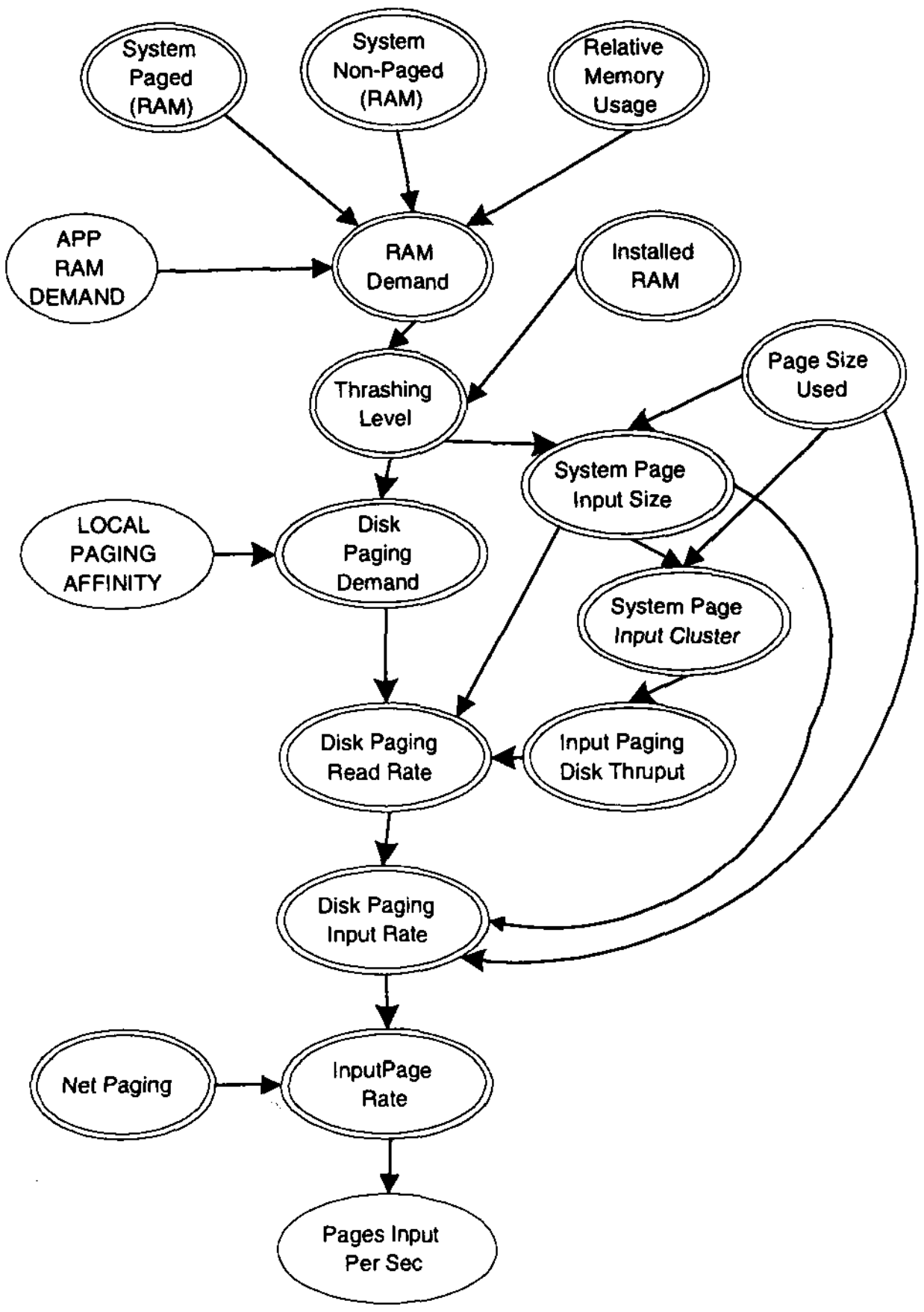

Figure 1: A fragment of the operating system functional model. All relationships are deterministic.

The model we have constructed addresses the domain of the server in a network, a system which provides services to other computers over network connections. This is a domain of particular interest to bottleneck detection since the responsiveness of the network is frequently limited by bottlenecks in the networks servers. We have deferred treatment of workstation graphics and the network as a whole.

The portion of the model which determines the amount of pressure on RAM page frames will illustrate its construction. A belief network of the model of paging behavior on Windows NT appears in Figure 1. The figure shows the structure of the model. The values of variables shown in double ovals are deterministic functions of the values of their predecessors. Deterministic nodes with no predecessors are constants based on the calibrated hardware or software, or are functions of variables in other portions of the model.

The application workload parameters are shown in Figure 1 as chance nodes (single ovals) at the upper left. The APP RAM DEMAND is the amount of RAM the application must access at steady state. The LOCAL PAGING AFFINITY is a number between 0 and 1 indicating the fraction of active virtual memory that is on the local server, as opposed to the fraction that is elsewhere on the network. At the bottom of Figure 1 is a chance node denoting a system counter or metric, Pages Input Per Second. This is the rate of input page traffic, a key system performance counter indicating how severely the system is thrashing or moving pages between RAM and disk or network. The other deterministic nodes in Figure 1 denote the internal operating system state variables that are well characterized when their predecessors are known, but are typically unmetered and hence unobservable.

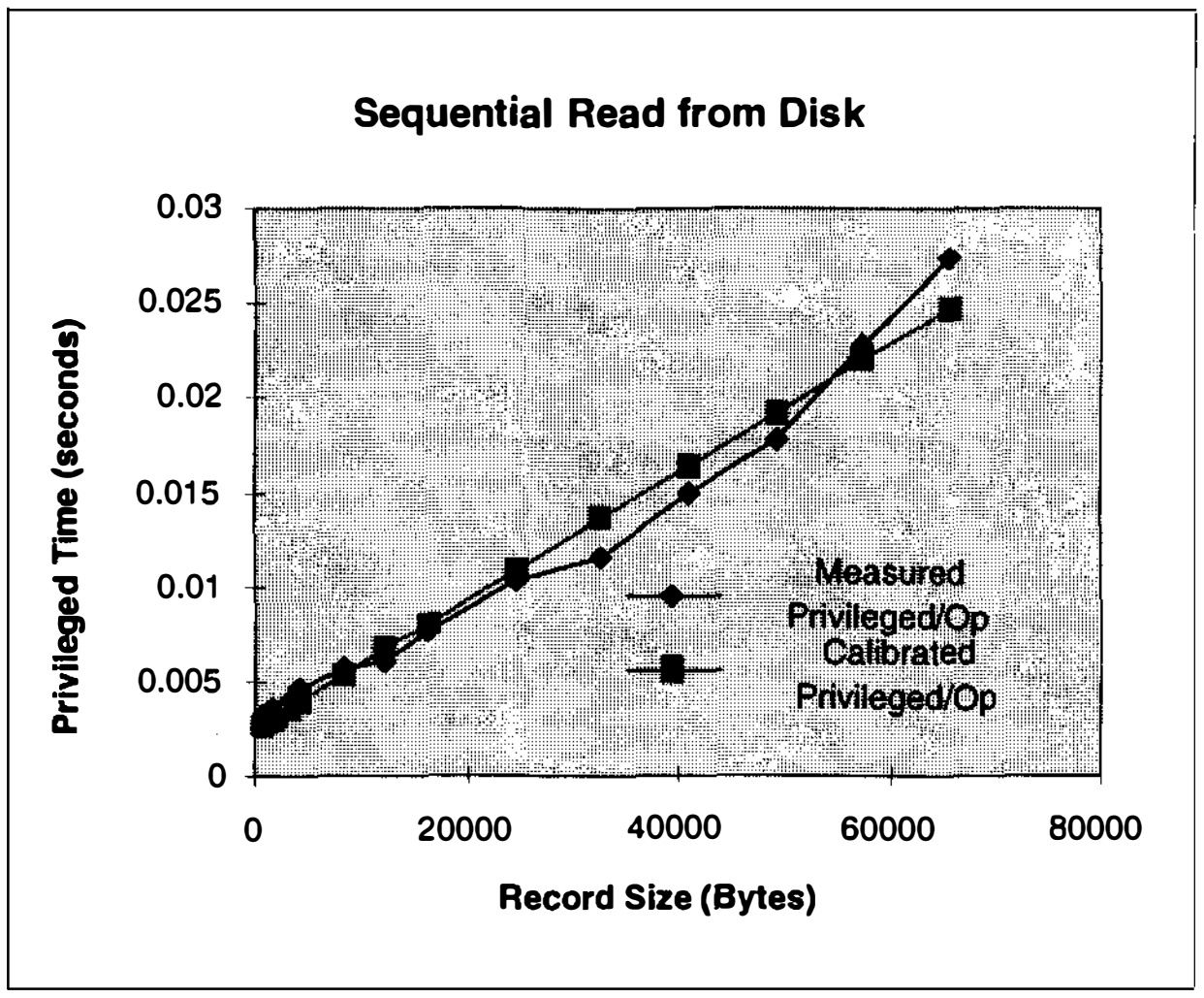

Figure 2: Calibration of sequential reads for a range of record sizes.

## 4.1 Calibration

In applying the model it is necessary to calibrate the hardware resource maximum bandwidths. This is ideally done on the system under test. Complex interactions between processor, memory, bus, controller, and device speeds make it difficult to measure the maximum throughput of a device in one system and extrapolate that throughput maximum to a different environment. Even identical processors with the same clock speed can deliver wildly different performance due to differences in size and design of the memory cache subsystem. Such differences can only be detected by calibration on a system of identical design and construction.

A synthetic workload generator is used to apply known workloads for calibration. The limits of throughput are collected for each resource over a range of key parameters, and placed into a data base for later extraction. When possible, simple linear regressions are performed to extract parameters for resource characteristics. Figure 2 illustrates one such regression. The processor overhead for disk operations is greatly dependent on the type of disk controller installed in the system. The fit is good, but deviations introduce additional uncertainty into the model.

## 4.2 Verification

Before the model can be put to use it is necessary to verify its accuracy. Let $\vec{w}$ be a vector of workload parameters of length $m$. Let $\vec{c}_p = f(\vec{w})$ be the vector of predicted counter values of length $n$. The synthetic workload generator is used to construct a series of one-dimensional workloads. Each generated workload exercises a single system service, varying a key workload parameter $w_i$, while holding the others fixed. Such a workload might be the sequential reading of a file from disk, with $w_i$ being the size of the record read, and taking on a sequence of increasing record sizes. These synthetic workloads are applied to a system, and the vector of actual performance counters, $\vec{c}_a$, are logged for each value assumed by the key workload parameter $w_i$.

An identical series of one-dimensional workloads is then applied to the model, and the predicted performance counters $\vec{c}_p$ are recorded for each level of $w_i$ . The results of the model's predictions are compared to the actual performance counters from the real system. Figure 3 illustrates a comparison of a particular actual counter $c_a^j$ to a corresponding model predicted counter value $c_p^j$ over a series of values assumed by a key workload parameter $w_i$.

## 4.3 Model Refinement

Verification results can be used to refine the model specification. For example we were concerned about the deviation in Figure 3 of model predicted processor utilization from that observed when the workload was applied to the real system. Examination of the atomic model intermediate values showed that the major component of processor demand was the processor being used by the operating system to read the data from the disk. In the model which produced Figure 3 this was expressed as a linear function of the application's read size in bytes.

A more refined model of this activity can be obtained by first regressing operating system processor usage against the size of a read from disk, as shown in Fig-ure 2. The operating system processor usage for reading from disk during sequential reads can then be determined from this regression by evaluating the regression formula at the size of the read-ahead record used by the operating system. The read-ahead size depends on the operating system kernel in Windows NT and not on the application record size. The operating system overhead for each read-ahead multiplied by the ratio of application read size to system read-ahead size gives the processor overhead for sequential reading.

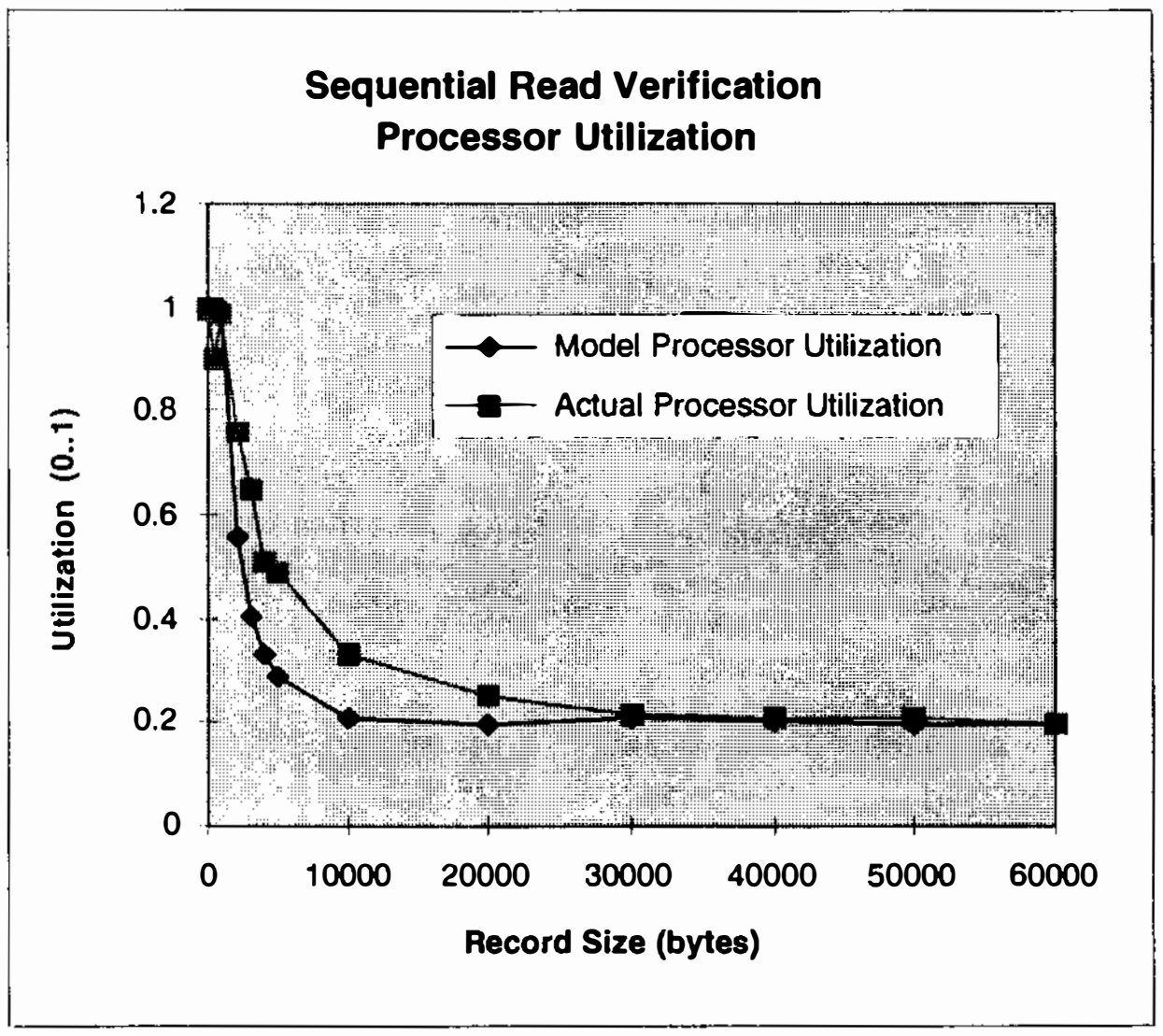

Figure 3: The initial verification for Sequential Reads

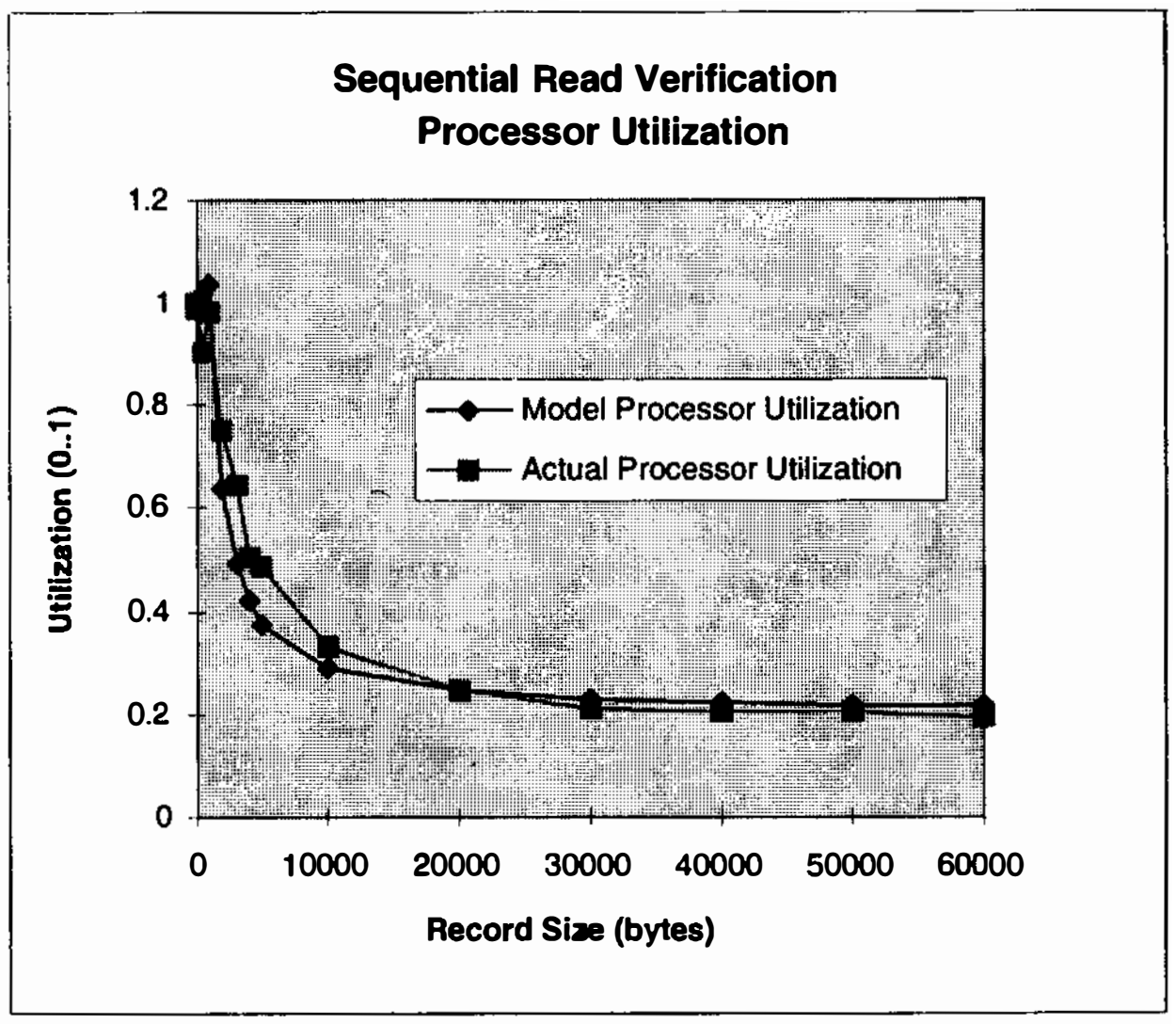

Figure 4: A revised verification for Sequential Reads

This new model produces the more accurate verification depicted in Figure 4. Although the refinement is an improvement over that shown in Figure 3, it is not perfect and the discrepancy is a continuing source of uncertainty.

One important initial question concerned the robustness of the model with respect to other computer systems than the one on which it is initially tested. As of this writing we have verified the model to acceptable levels across a number of systems, including an Intel 486, an Intel Pentium, and a DEC Alpha, all running Windows NT.

## 5 Inference

During inference we wish to determine those values of the workload parameters that best explain the observed performance counter values. As before, $\vec{w}$ is a vector of workload parameters of length $m$ and $\vec{c}_p = f(\vec{w})$ is the vector of predicted counter values of length $n$. As before, let $\vec{c}_a$ be the vector of actual counter values corresponding to $\vec{c}_p$. The function $f$ captures the dependence of the counter values on the workload and has been described in Section 4. In this section we describe two methods for reasoning with this model, that is finding the value of $\vec{w}$ that best explains $\vec{c}_a$.

### 5.1 Inversion

For an initial formulation of this problem we choose to not represent uncertainty explicitly. We let $\vec{c}_p = \vec{c}_a$ and then the best explanation is obtained by solving for workload parameters:

$$\vec{w}^* = f^{-1}(\vec{c}_a)$$

Unfortunately, we cannot solve for $\vec{w}$ analytically due to numerous discontinuities and non-linearities in the model $f$. The discontinuities arise in discrete shifts in operating system algorithms, such as differences in file system implementation when the record size is a multiple of the page size. Thus, in applying this technique we use numerical methods to search for the $\vec{w}^*$ that is a solution to:

$$\min_{\vec{w}} \sum_{i=1}^{n} \alpha_i (c_a^i - f^i(\vec{w}))^2 \qquad (1)$$

where the $\alpha_i$ are weighting or normalization factors. We refer to this inference method as *inversion*.

We have used a Monte Carlo technique to explore the solution surface and provide evidence that there is a single solution for a variety of workloads, searching over a substantial number of samples.

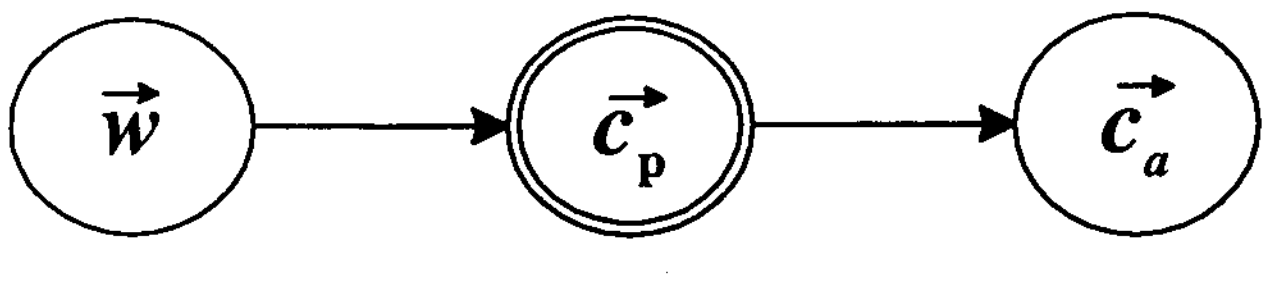

Figure 5: Belief network structure for operating system model.

### 5.2 Most Probable Explanation

From a probabilistic perspective, the best explanation of the observed counters is that workload assignment with maximum probability given the data, that is the assignment that is the most probable explanation (MPE) for the observations.

The uncertainty structure for this problem is shown in Figure 5 as a belief network. The network represents the conditional independencies we have asserted in this domain [Pearl, 1988]. The functional relationship between workload ($\vec{w}$) and predicted counters ($\vec{c}_p$), the atomic model, is reflected in the double-oval representation in Figure 5.

Although the inversion method has been effective in initial experiments in identifying bottlenecks for pure workloads, it ignores several critical factors. It is non-Bayesian to the extent that it disregards prior information regarding the distribution of application workloads. Also, it does not explicitly account for modeling and measurement uncertainty in the relationship between actual and predicted counters. For example the fact that the measurement system is running introduces some systematic biases into the observed counters that are not explicitly modeled. The probabilistic inference measure accounts for these issues implicitly by modeling the error in the predictions.

Uncertainty in the belief network is characterized by $\Pr(\vec{w}|\xi)$, the prior distribution of workload parameters, and $\Pr(\vec{c}_a|\vec{c}_p,\xi)$ the uncertain relationship between the predicted and actual counter values. In these expressions, $\xi$ is background information including such factors as the installed hardware and the version of the operating system software[1].

The MPE assignment $\vec{w}^*$ is that set of workload parameters $\vec{w}$ that has the maximum probability given the observed counters, that is

$$\Pr(\vec{w}^*|\vec{c}_a,\xi) = \max_{\vec{w}} \Pr(\vec{w}|\vec{c}_a,\xi) \quad (2)$$

Given the belief network model, we can write the joint probability of the variables of interest as:

$$\begin{aligned}\Pr(\vec{w},\vec{c}_p,\vec{c}_a|\xi) &= \Pr(\vec{c}_a|\vec{c}_p,\xi)\Pr(\vec{c}_p|\vec{w},\xi)\Pr(\vec{w}|\xi)\\ &= \Pr(\vec{c}_a|f(\vec{w}),\xi)\Pr(\vec{w}|\xi)\end{aligned}$$

and we have

$$\Pr(\vec{w}|\vec{c}_a,\xi) = \frac{\Pr(\vec{c}_a|f(\vec{w}),\xi)\Pr(\vec{w}|\xi)}{\Pr(\vec{c}_a|\xi)}$$

Since the denominator is a constant in any particular case, Equation 2 becomes:

$$\Pr(\vec{w}^*|\vec{c}_a,\xi) = k \max_{\vec{w}} \Pr(\vec{c}_a|f(\vec{w}),\xi)\Pr(\vec{w}|\xi) \quad (3)$$

where $k$ is a constant.

In evaluating this expression, we made two sets of assumptions. First, we assumed that the workload parameters are marginally independent, as indicated in Figure 5. In various experiments, we have assumed these parameters to be either uniformly or lognormally distributed. With lognormally distributed workloads we have:

$$\Pr(w_i,\xi) = (\sigma_i(w_i - a_i)\sqrt{2\pi})^{-1} e^{-(\ln(w_i-a_i)-\mu_i)^2/2\sigma_i^2} \quad (4)$$

where $\mu_i$, $\sigma_i$, and $a_i$ are the logarithmic mean, standard deviation, and minimum value respectively for workload component $w_i, i = 1\ldots m$. The values of these parameters are provided by direct assessment from an expert.

Second, we assume a multivariate Gaussian error model, that is $\vec{c}_a = f(\vec{w}) + \vec{\epsilon}$ where $\vec{\epsilon} \sim N(\vec{\mu}_\epsilon,\Sigma)$ and $\vec{\mu}_\epsilon$ is the vector of mean errors and $\Sigma$ is an $n$ by $n$ covariance matrix [DeGroot, 1970]. We estimate the mean errors and covariance from a sample of known workloads, model predictions, and actual counter values on the target system. Using techniques from [DeGroot, 1970], we can update the parameters of the error model by assuming that the distribution for $\vec{\mu}_\epsilon$ is multivariate normal and the distribution for $\Sigma^{-1}$ is Wishart. For purposes of this study, we will estimate $\vec{\mu}_\epsilon$ using the sample mean error and estimate $\Sigma$ with the sample covariance from a set of verification samples.

Using procedures similar to those applied during verification, we can run a set of controlled experiments on the target machine to generate model error data. For a set of sampled known workloads, we generate model predictions (using the calibrated model) and collect actual counter values. This is the sample for estimating the covariance matrix.

[1] Recall that model parameters relating to inherent hardware and software speed on a particular machine and release of the operating system are fixed during calibration.

| App Workload | Counters |
|---|---|
| *Inter-operation cpu times:* | System.PctPriv |
| -Sequential Write | System.PctUser |
| -Sequential Read | System.SystemCallRate |
| -Random Read | Disk.DiskReadByteRate |
| -Random Write | Disk.DiskReadRate |
| Sequential Read Size | Disk.DiskWriteByteRate |
| Sequential Write Size | Disk.DiskWriteRate |
| Random Read Size | Cache.CopyReadHitsPct |
| Random Write Size | Cache.CopyReadsPerSec |
| Random Read Extent | Cache.LazyWritePgsPerSec |
| Random Write Extent | Memory.PgFaultsPerSec |
| RAM Demand | Memory.CacheFaultsPerSec |
| | Memory.PagesInputPerSec |
| | Memory.PagesOutputPerSec |

Table 1: Workloads and counters that are incorporated into the current model.

Since the actual error given a set of workload parameters is just $\vec{c}_a - f(\vec{w})$, then the probability of the observed counter values, given the model $f$ and $\vec{w}$ is calculated as follows.

$$\Pr(\vec{c}_a | f(\vec{w}), \xi) = ce^{-1/2(\vec{c}_a - f(\vec{w}) - \vec{\mu}_\epsilon)^T \Sigma^{-1} (\vec{c}_a - f(\vec{w}) - \vec{\mu}_\epsilon)} \quad (5)$$

where $c = (2\pi)^{-n/2}|\Sigma|^{-1/2}$. Equations 4 and 5 are used to evaluate Equation 3.

As discussed previously, the function $f$ is not well behaved since it incorporates various thresholding behaviors and integer constraints in the operating system. Therefore these optimization problems cannot be performed analytically and we have used numerical search techniques to find solutions. We discuss these results in the following section.

## 6 Implementation and Results

The Windows NT system model and inference procedure have been implemented in Microsoft Excel. We utilize the Excel Solver feature to provide the numerical optimization procedure to search for the desired workload vector $\vec{w}$ under the inversion method (Equation 1) as well as the MPE assignment (Equation 3). The application workload parameters ($\vec{w}$) and counters ($\vec{c}_a, \vec{c}_p$) that are in the model are listed in Table 1.

In order to test the inference procedures, we impose a set of known workload parameters on a given platform and collect actual performance monitor counters. The model is then used to infer the workloads. Our test suite consists of the following cases which vary as to the nature of the bottleneck and its cause:

- Sequential Read
- Sequential Write
- Random Read
- Random Write
- Paging

In each of these cases, the both the inversion and MPE methods are able to identify the correct bottleneck. For the MPE method, we have been able to estimate the error model on a verification set of over 180 cases representing model predictions and actual values for three different machine configurations. Even with this sparse dataset, performance is encouraging. The Figure 6 shows that for every scenario, the model is finding the correct bottleneck and a good approximation of demand.

In Figure 7, we see additional detail regarding utilizations for the case of Sequential Reads. Again the actual levels of each variable along with inferred values using the MPE and inversion methods are shown. Again, both methods are finding the correct solution.

To this point in development, both methods have been able to correctly identify pure workloads and bottlenecks. We anticipate that for more diverse, real-world loads, the MPE method using an error model estimated on the target machine will provide the best performance. In addition, in the cases we studied, the MPE method converges to the solution much faster than inversion. This is probably due to the improved directional information provided by using a full covariance matrix.

## 7 Conclusions and Future Work

We have developed a core model and inference methodology for detecting computer system bottlenecks. We are currently extending and verifying the model in network and graphics subsystems. The probabilistic inference methodology relies on a substantial amount of data during verification to learn the error model. Fortunately, in the realm of computer performance analysis it is relatively easy to generate the needed data and we are working to automate that data collection effort.

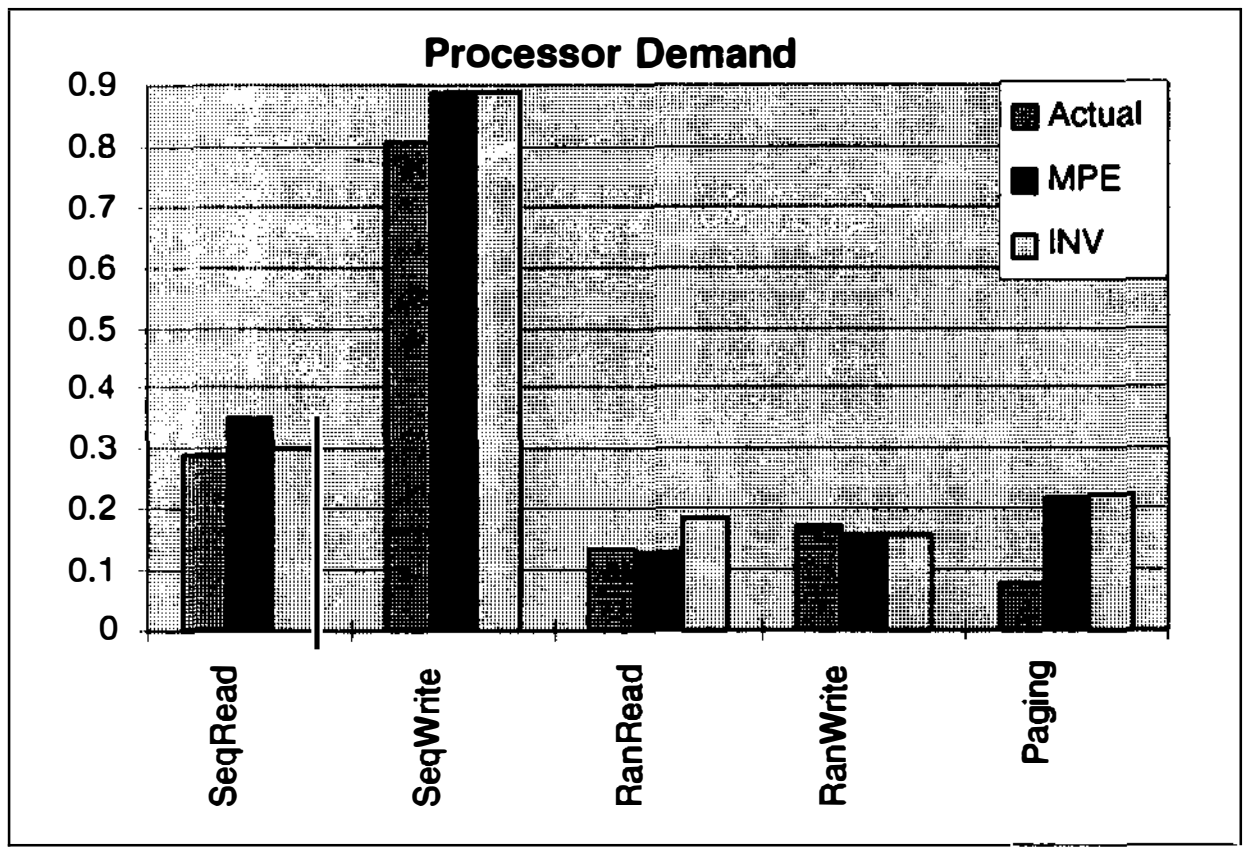

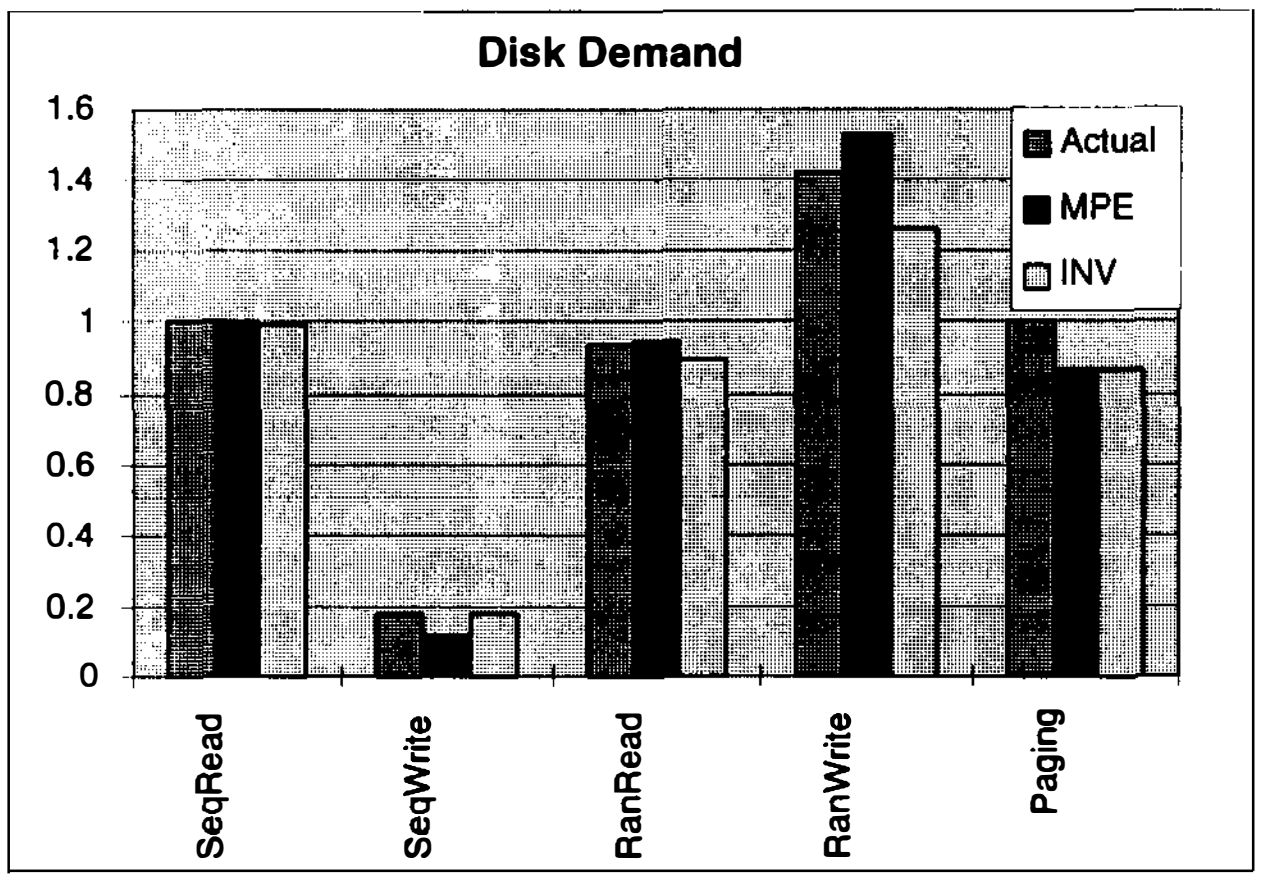

Figure 6: Comparison actual versus inferred processor and disk demands for 5 scenarios using the inversion method and MPE inference method.

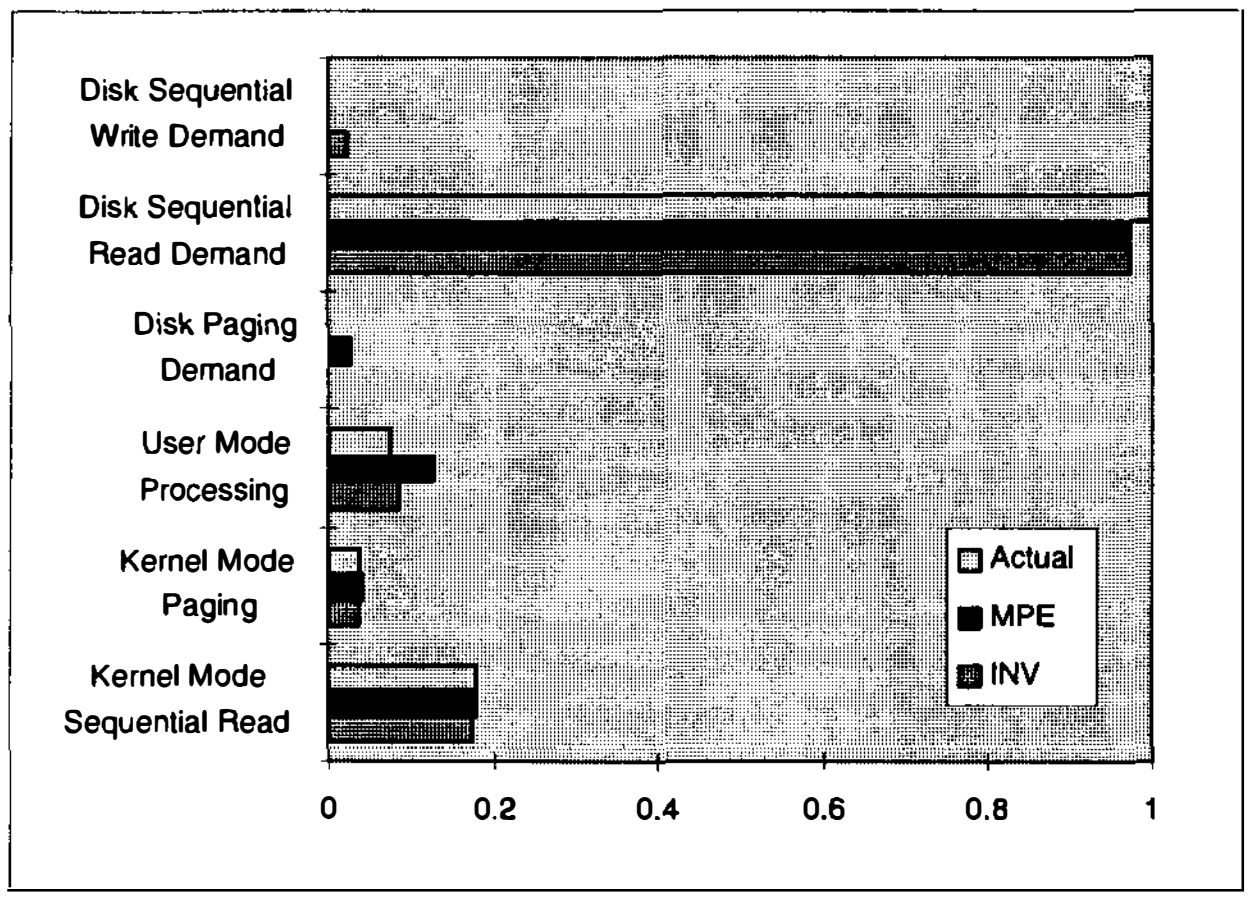

Figure 7: Comparison actual versus inferred device utilizations for Sequential Read using the inversion method and MPE inference method.

We have presented this material primarily in terms of detecting bottlenecks, a diagnostic task. However, as in medicine, diagnosis is rarely an end in itself—ultimately we wish to "treat" the patient. In the computer domain, the primary classes of therapy are changing hardware (e.g. adding memory to improve performance) and redesigning software to utilize less critical resources.

We plan to use similar approaches to predict the effects of changes to application workload parameters. The model can predict throughput and bottlenecks given an increment to application workloads. It also can be used by software developers to predict the performance of their application, and to help determine which portions of the program merit additional design and implementation effort.

Finally, since the model includes many variables relating to operating system design and algorithms, this approach can address issues relating to the structure of the operating system itself. This would include off-line design studies, for example, estimating the possible system-wide effects of different paging algorithms. Similar models could also potentially be used for dynamic tuning of system operating parameters, such as cache sizes, in response to inferred application loadings.

## Acknowledgments

The authors thank David Heckerman and the anonymous referees for useful comments and suggestions.